\documentclass[9pt,twocolumn,twoside]{pnas-new}

\usepackage{graphicx}			
\usepackage{subfigure}
\usepackage{float}
\usepackage{amsmath,amssymb,bm}
\usepackage{mathtools}

\templatetype{pnasresearcharticle} 

\addto\captionsenglish{}
\addto\captionsenglish{}

\title{Classificação de espécies de peixe utilizando redes neurais convolucional}

\author[a,1]{Andre G. C. Pacheco}

\affil[a]{Programa de Pós-Graduação em Informática (PPGI), Universidade Federal do Espírito Santo (UFES), Vitória - ES, Brasil}

\dates{Este relatório técnico foi produzido em Novembro de 2016}

\doi{\url{Relatório Técnico}}

\leadauthor{André G. C. Pacheco} 

\correspondingauthor{\textsuperscript{1}Autor correspondente. E-mail: agcpacheco@inf.ufes.br}


\begin{abstract}

Classificação de dados está presente em diversos problemas reais, tais como: reconhecer padrões em imagens, diferenciar peças defeituosas em uma linha de produção, classificar tumores benignos e malignos, dentre diversas outras. Muitos desses problemas possuem padrões de dados difíceis de serem identificados, o que requer, consequentemente, técnicas mais avançadas para sua resolução. Recentemente, diversos trabalhos abordando diferentes arquiteturas de redes neurais artificiais vêm sendo aplicados para solucionar problemas de classificação. Quando a classificação do problema deve ser obtida por meio de imagens, atualmente a metodologia padrão é uso de redes neurais convolucionais. Sendo assim, neste trabalho são utilizadas redes neurais convolucionais para classificação de espécies de peixes. 

\end{abstract}

\begin{document}

\maketitle
\thispagestyle{firststyle}
\ifthenelse{\boolean{shortarticle}}{\ifthenelse{\boolean{singlecolumn}}{\abscontentformatted}{\abscontent}}{}

\section{Introdução}
As Redes Neurais Artificiais (RNA) se tornaram populares no final da década de 1980 com o surgimento e aperfeiçoamento do algoritmo de treinamento \textit{backpropagation} \cite{williams1986}. Desde então, as	RNA's vêm sendo aplicadas em diversas áreas da computação, tais como: robótica \cite{mohareri2012}, tratamento de imagens \cite{d2013}, predição de séries temporais \cite{hrasko2015}, classificação de dados \cite{krizhevsky2012}, dentre muitas outras áreas. Ao longo do tempo, as teorias desenvolvidas com base em RNA's direcionaram ao desenvolvimento de uma nova abordagem, com uma maior capacidade de aprendizagem e generalização, para problemas nos quais a natureza dos padrões são de difícil compreensão. Essa nova abordagem é conhecida como aprendizado profundo, que apresenta-se como uma melhor alternativa para problemas que envolvam comportamento inteligente, como classificação de padrões \cite{bengio2009}.

No contexto de aprendizagem profunda, pode-se destacar as redes profundas de crença (DBNs) \cite{hinton2006}, obtidas, por exemplo, por meio de empilhamento de máquinas de Boltzmann restrita (RBMs) \cite{hinton2002}, e as redes neurais convolucionais (CNN) \cite{lecun1998}. Ambas arquiteturas utilizam diversas camadas de neurônios ocultos com intuito de extrair características de um dado conjunto de dados. Atualmente, as duas abordagens vêm sendo utilizadas em diversos trabalhos nas mais diversas áreas.

CNNs foram primeiramente propostas em 1998 por LeCun et al. \cite{lecun1998}, na qual os autores desenvolveram uma arquitetura neural conhecida como LeNet5, utilizada para reconhecer dígitos escritos à mão. Na ocasião, tal arquitetura estabeleceu um novo estado da arte ao atingir 99.2\% de acurácia na base de dados MNIST \cite{mnist}. Anos mais tarde, a Krizhevsky et. al \cite{krizhevsky2012} estabeleceram um marco na área ao propor a AlexNet, arquitetura vencedora do desafio ImageNet \cite{deng2009}. Desde então, aplicação mais popular relacionada a CNNs vêm sendo reconhecer padrões em imagens, todavia, diversos trabalhos vêm aplicando CNNs em outros tipos de tarefas \cite{wallach2015, gibney2016}. 

Neste trabalho são utilizadas diferentes arquiteturas de CNNs para classificação de espécie de peixes. Tal base de dados é composta por milhares de imagens que pode conter um, mais de um ou nenhum peixe, sempre da mesma espécie. Este problema de classificação faz parte do desafio \textit{The Nature Conservancy Fisheries Monitoring}, hospedado da plataforma \textit{Kaggle} \cite{kaggle}. O restante deste trabalho esta organizado da seguinte forma: na seção 2 são apresentados os conceitos básicos relacionados a CNNs; na sequência, na seção 3 a base de dados e o desafio são apresentados; em seguida, na seção 4, os resultados experimentais são descritos; por fim, na seção 5 são apresentadas as conclusões do trabalho.

\section{Redes Neurais Convolucionais}
Redes neurais convolucionais são similares a redes neurais tradicionais: ambas são compostas por neurônios que possuem pesos e \textit{bias} que necessitam ser treinados. Cada neurônio recebe algumas entradas, aplica o produto escalar das entradas e pesos além de uma função não-linear. Ademais, ambas possuem a última camada toda conectada e todos os artifícios utilizados para melhorar a rede neural tradicional também são aplicados nesta camada. Dessa forma, qual a vantagem de se utilizar uma CNN? Uma CNN assume que todas as entradas são imagens, o que permite codificar algumas propriedade na arquitetura. Redes neurais tradicionais não são escaláveis para imagens, uma vez que a mesma produz um número muito alto de pesos a serem treinados \cite{li2015}.

\begin{figure}[h]
\centering
\includegraphics[scale=0.2]{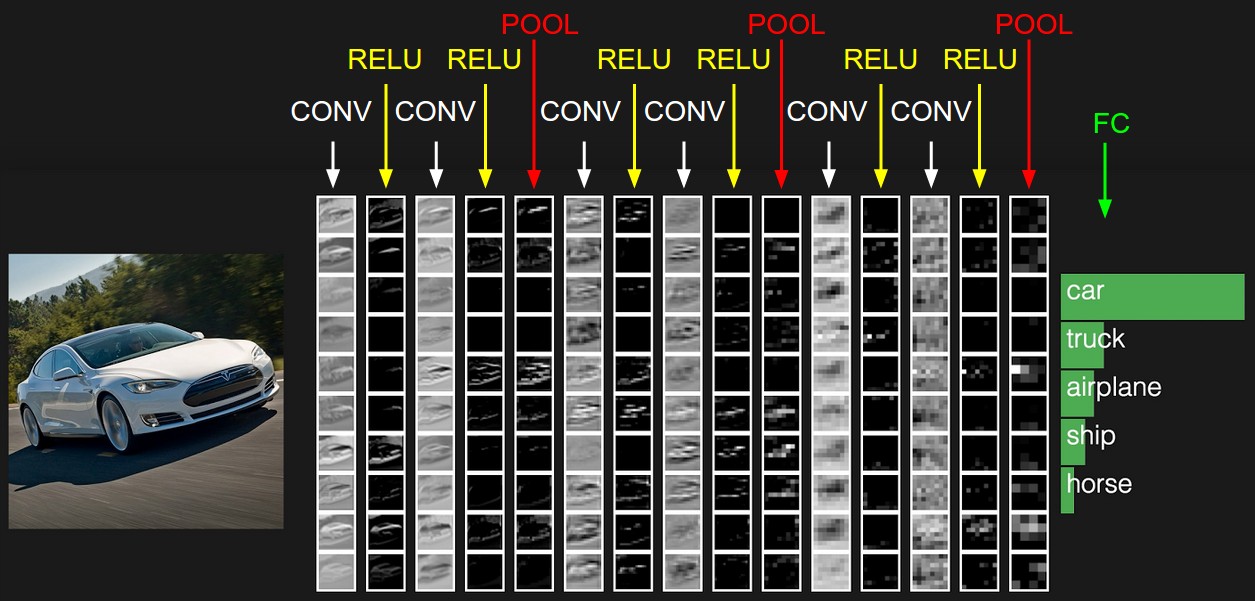}
\caption{Exemplo de arquitetura de uma CNN}
\label{fig:arqCnn}
\end{figure}

Uma CNN é composta por uma sequência de camadas. Além da camada de entrada, que normalmente é composta por uma imagem com largura, altura e profundidade (RGB), exitem três camadas principais: camada convolucional, camada de \textit{pooling} e camada totalmente conectada. Além disso, após uma camada de convolução é comum uma camada de ativação (normalmente uma função ReLu). Essas camadas, quando colocadas em sequência (ou empilhadas), forma uma arquitetura de uma CNN, como ilustrada na Figura \ref{fig:arqCnn}. Na sequência as camadas principais serão descritas, bem como suas funções.

\subsection{Camada convolucional} 
A camada convolucional é a camada mais importante da rede. Nela é realizada a parte mais pesada do processamento computacional. Esta camada é composta por um conjunto de filtros (ou \textit{kernels}) capazes de aprender de acordo com um treinamento. Os filtros são matrizes pequenas (por exemplo, $5 \times 5 \times 3$) compostas por valores reais que podem ser interpretado como pesos. Esses filtros são convoluídos com os dados de entradas para obter um mapa de características. Estes mapas indicam regiões na qual características específicas em relação ao filtro, são encontradas na entrada. Os valores reais dos filtros se alteram ao longo do treinamento (assim como os pesos de uma rede neural tradicional) fazendo com que a rede aprenda a identificar regiões significantes para extrair características do conjunto de dados.

\begin{figure}[h]
\centering
\includegraphics[scale=0.2]{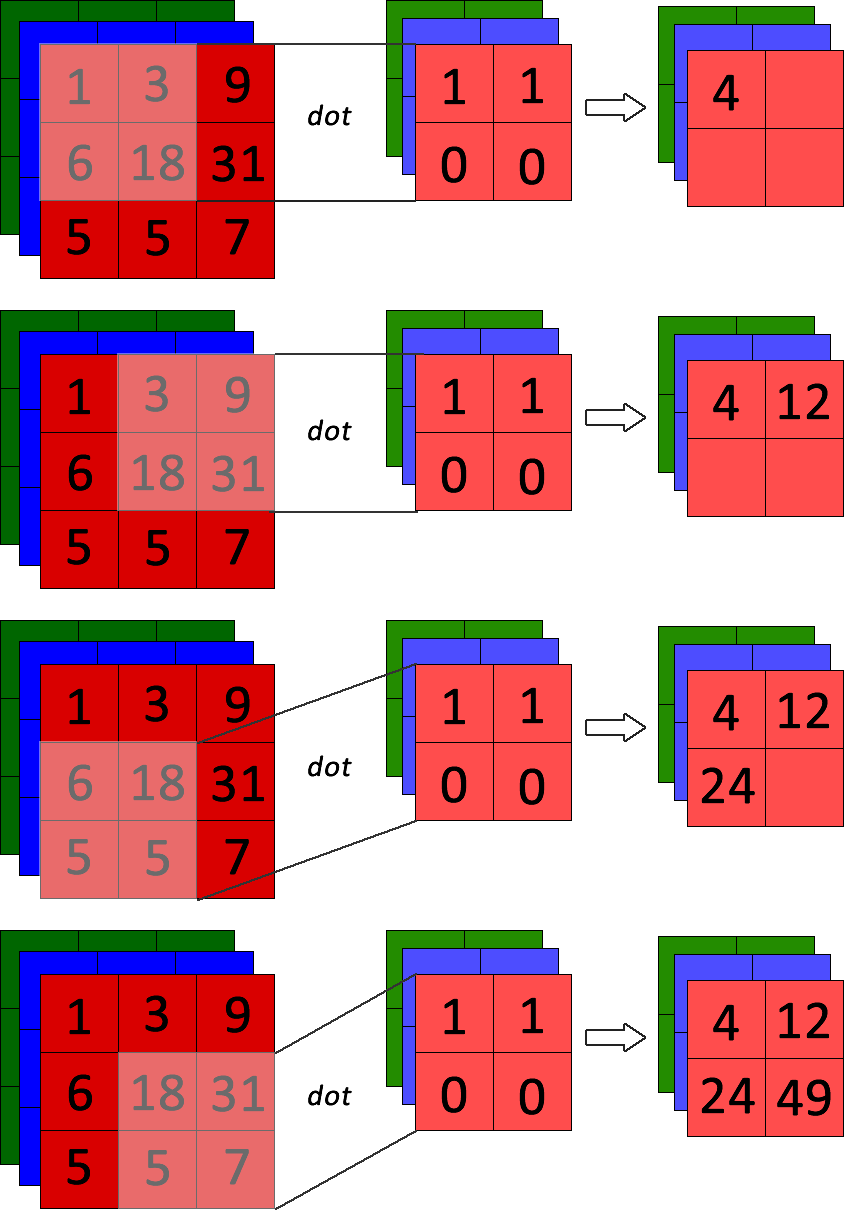}
\caption{Exemplo de convolução entre a imagem e um filtro}
\label{fig:exconv}
\end{figure}

A convolução entre um filtro e a imagem é ilustrado na Figura \ref{fig:exconv}. Neste exemplo é apresentada uma imagem $4 \times 4 \times 3$ e um filtro $2 \times 2 \times 3$. A convolução é realizada através do produto escalar de uma região da imagem do tamanho do filtro pelo filtro propriamente dito. Na sequência o filtro é deslizado para outra região e o produto escalar é realizado novamente até que toda a imagem seja percorrida. Perceba que cada canal é convoluído por uma dimensão diferente do filtro, que neste exemplo esta representado pela mesma cor do canal (RGB). Neste exemplo o filtro desliza na imagem deslocando 1 pixel para lado ou para baixo. Este deslizamento é controlado pelo parâmetro conhecido como \textit{stride}, que neste caso é 1. Este valor pode ser alterado, todavia deve respeitar os limites da imagem. Caso o tamanho do filtro e do \textit{stride} desejado não seja possível para a imagem, pode-se utilizar \textit{zero-padding}, que nada mais é do que adicionar zeros na borda da imagem para tornar o deslizamento do filtro possível. Tanto o \textit{stride} quanto \textit{zero-padding} são parâmetros de projetos que deve ser analisados pelo projetista. Por fim, vale a pena destacar que podem existir (e normalmente existem) mais de um filtro por camada de convolução. Dessa forma, cada filtro resulta em uma saída de três dimensões, como o da Figura \ref{fig:exconv}.

Nas matrizes resultados da convolução é aplicado a função de ativação. A função mais utilizada é a ReLU (unidade de retificação linear), que é simplesmente aplicar a função $max(0,x)$ em cada elemento do resultado da convolução. No exemplo da Figura \ref{fig:exconv}, todos os elementos são maiores do que zero, sendo assim, o resultado da ReLU são os próprios valores. Por fim, como os pesos utilizados por cada filtro são os mesmos em todas as regiões, ele são conhecidos como pesos compartilhados. Essa característica reduz consideravelmente a quantidade de pesos da CNN quando comparadas a uma RNA.

\subsection{Camada de \textit{pooling}}
\begin{figure}[h]
\centering
\includegraphics[scale=0.25]{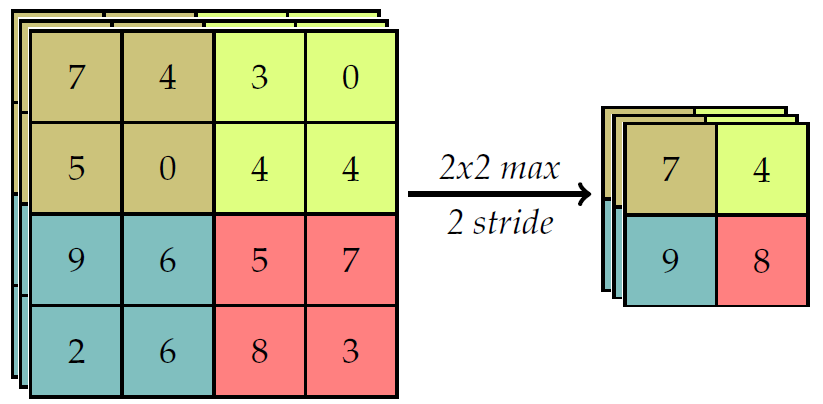}
\caption{Exemplo de \textit{maxpooling} de uma imagem $4 \times 4$}
\label{fig:pooling}
\end{figure}

É muito comum após uma camada de convolução existir uma camada de \textit{pooling}. A técnica de \textit{pooling} é utilizada com objetivo de reduzir o tamanho espacial das matrizes resultantes da convolução. Consequentemente, essa técnica reduz a quantidade de parâmetros a serem aprendidos na rede, contribuindo para o controle de \textit{overfitting}. As camadas de \textit{pooling} operam de maneira independente em cada um dos canais do resultado da convolução. Além disso, é necessário determinar previamente o tamanho do filtro e \textit{stride} para realizar o \textit{pooling}. Na Figura \ref{fig:pooling} é ilustrado a realização de \textit{pooling} considerando que o resultado da convolução seja uma matriz $4 \times 4 \times 3$, o filtro do \textit{pooling} abrange regiões $2 \times 2$ em cada canal, com \textit{stride} igual a 2. Além disso, a função de agregação é $max (X)$, ou seja, escolhe o maior valor da região.

\noindent Como pode ser observado, a camada de \textit{pooling} não reduz a quantidade de canais e sim a quantidade de elementos em cada canal. Além do \textit{maxpooling}, pode ser utilizadas outras funções, como por exemplo a média dos valores. Todavia, a função \textit{max} vem obtendo melhores resultados, uma vez que pixels vizinhos são altamente correlacionados.

\subsection{Camada totalmente conectada}
Diferentemente da camada convolucional, na qual os pesos são conectados apenas em uma região, a camada totalmente conectada, como o próprio nome já sugere, é completamente conectada com a camada anterior. Tipicamente são utilizadas como última camada da CNN e funciona da mesma maneira que as redes neurais tradicionais. Portanto, os mesmos artifícios para melhorar o desempenho de uma RNA, como \textit{dropout} \cite{srivastava2014} por exemplo, também são aplicáveis nesta camada. 

\begin{figure}[h]
\centering
\includegraphics[scale=0.50]{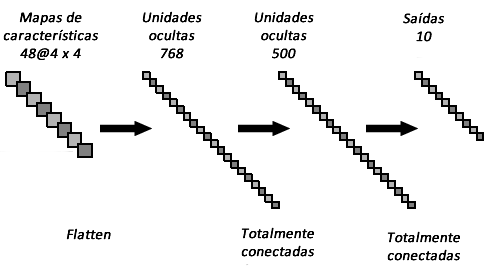}
\caption{Ilustração de uma camada totalmente conectada}
\label{fig:fc}
\end{figure}

Como a camada totalmente conectada vêm após uma camada convolucional ou de \textit{pooling}, é necessário conectar cada elemento das matrizes de saída de convolução em um neurônio de entrada. A Figura \ref{fig:fc} ilustra a conexão de uma camada convolucional com uma camada totalmente conectada. É possível observar que os 48 mapas de características $4 \times 4$ são colocados de forma linear formando 768 entradas para camada totalmente conectada, que por sua vez, possui 500 neurônios ocultos que resultam em 10 saídas para rede. Neste ponto, é possível observar visualmente um rede neural tradicional.

\noindent Nas unidades de saída, que no exemplo da Figura \ref{fig:fc} são 10, é utilizado uma função \textit{softmax} para se obter a probabilidade de dada entrada pertencer a uma classe. Neste ponto é realizado o algoritmo de treinamento supervisionado \textit{backpropagation}, assim como em uma rede neural tradicional. O erro obtido nesta camada é propagado para que os pesos dos filtros das camadas convolucionais sejam ajustados \cite{lecun1998}. Dessa forma, os valores dos pesos compartilhados são aprendidos ao longo do treinamento.

\subsection{Arquitetura de uma CNN}
A arquitetura de uma CNN é relacionada com a maneira que as camadas descritas anteriormente são organizadas. De fato o projetista pode organizá-las da maneira que julgar adequado. A maior parte das arquiteturas utilizam uma camada de entrada, obviamente, seguida de um bloco de $N$ camadas convolucionais com ativação ReLU conectada a uma camada de \textit{pooling}. Esse bloco é repetido $M$ vezes ao longo da rede que ao final é conectado a uma camada totalmente conectada para determinar a classificação final. Além disso, na montagem da arquitetura também são definidos parâmetros como tamanho do filtro, \textit{stride} e \textit{zero-padding}. Todas essas escolhas impactam diretamente na quantidade de pesos que a rede deve treinar e por consequência o quanto de computação é necessário para que esses valores sejam bem ajustados.

Existem diversas arquiteturas bem conhecidas na área de redes convolucionais. Três das mais utilizadas são descritas na sequência:

\begin{itemize}
\item \textbf{LeNet-5:} desenvolvida por LeCun et al. \cite{lecun1998} foi a primeira aplicação de CNN que obteve sucesso. Foi utilizada para classificar dígitos, por conta disso, muito utilizada em tarefas como leitura de códigos postais e afins. A Figura \ref{fig:lenet} ilustra a arquitetura de uma LeNet-5, que possui uma camada de entrada com imagens de $32 \times 32$ \textit{pixels} e mais 7 camadas de pesos treináveis. As camadas convolucionais utilizam filtros $5 \times 5 \times 3$ e $6 \times 6 \times 3$ para computar os mapas de características e filtros $2 \times 2 \times 3$ para realização de \textit{pooling}. Por fim, além da camada de saída que possui conexão para 10 classes, a rede possui duas camadas totalmente conectadas de 120 e 84 neurônios.

\begin{figure}[h]
\centering
\includegraphics[scale=0.38]{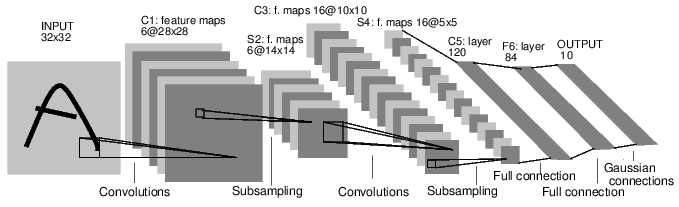}
\caption{Arquitetura de uma LeNet-5, CNN utilizada para reconhecimento de dígitos}
\label{fig:lenet}
\end{figure}

\item \textbf{AlexNet:} a arquitetura AlexNet foi o trabalhou que popularizou a CNNs para visão computacional. Proposta por Krizhevsky et al. \cite{krizhevsky2012}, como já mencionado anteriormente, foi a metodologia vencedora do desafio ImageNet em 2012 \cite{deng2009}. Sua arquitetura é bastante similar à da LeNet-5, porém ela é mais profunda, com mais camadas convolucionais, e possui muito mais mapas de característica, como mostrado na Figura \ref{fig:alexnet}. A AlexNet recebe como entrada imagens de $224 \times 224$ \textit{pixels} por canal. Na primeira camada de convolução utiliza um filtro de $11 \times 11 \times 3$, na segunda $5 \times 5 \times 3$ e na terceira a diante $3 \times 3 \times 3$. Além disso, a terceira, quarta e quinta camada são conectadas sem utilização de \textit{pooling}. Por fim, a a rede possui duas camadas totalmente conectadas com 2048 neurônios cada e uma camada de saída com 1000 neurônios, quantidade de classes existentes no problema. Vale a pena destacar que a AlexNet foi a primeira rede a utilizar \textit{dropout} para auxiliar no treinamento da camada totalmente conectada. Além disso, o treinamento de toda a rede foi realizado utilizando duas GPUs. Como pode ser observado na Figura \ref{fig:alexnet}, a rede é dividida em duas partes, uma GPU executou a parte de cima e a outra a parte de baixo da rede. 

\begin{figure}[h]
\centering
\includegraphics[scale=0.23]{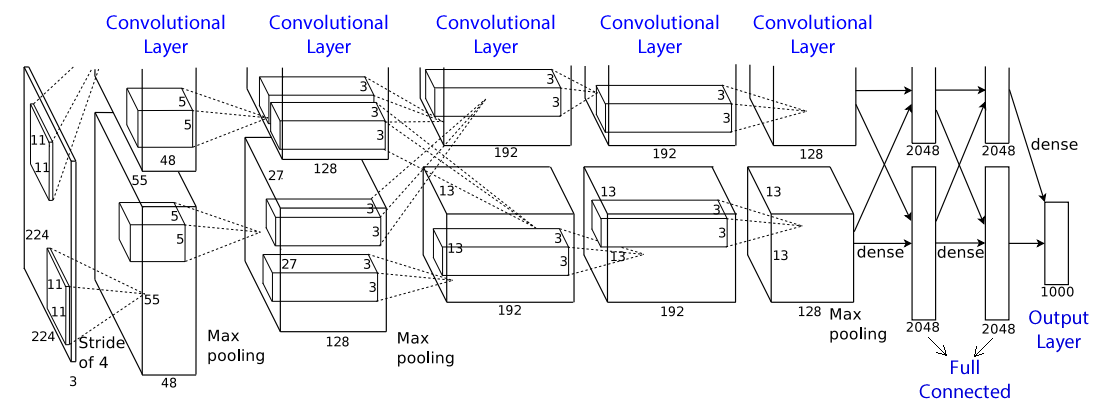}
\caption{Arquitetura de uma AlexNet, CNN vencedora do desafio ImageNet em 2012}
\label{fig:alexnet}
\end{figure}

\item \textbf{VGGNet:} a arquitetura VGGNet foi proposta por Simonyan \& Zisserman \cite{simonyan2014} e teve como principal contribuição mostrar que a profundidade da rede é um componente crítico para uma boa performance. O modelo padrão da VGGNet possui diversas camadas de convolução aplicando filtros $3 \times 3 \times 3$ e \textit{maxpooling} com filtros $2 \times 2 \times 3$. Na Figuro \ref{fig:vgg} é ilustrado uma comparação entre uma AlexNet e uma VGGNet. É possível observar uma camada de \textit{pooling} sempre após duas de convolução. Além disso, a rede possui três camadas totalmente conectadas além da camada de saída. Devido a sua profundidade, a VGGNet é bem cara computacionamente e necessita de muita memória para computar em cima de seus parâmetros (140M).

\begin{figure}[h]
\centering
\includegraphics[scale=0.25]{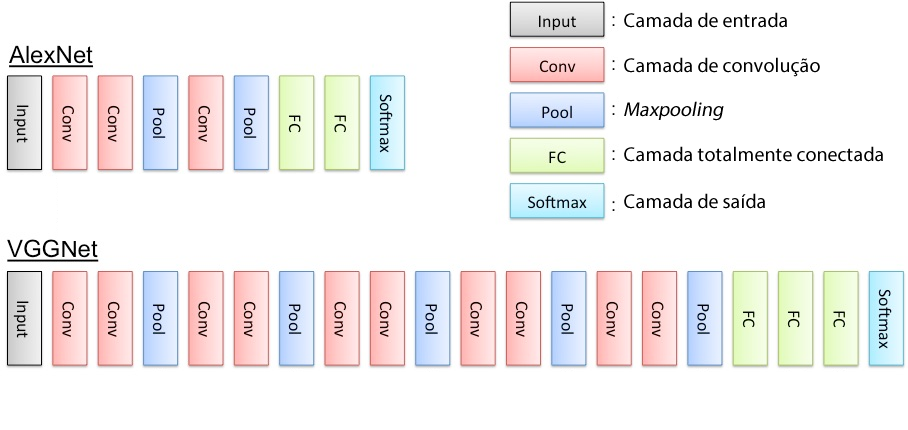}
\caption{Comparação entre as arquiteturas AlexNet e VGGnet}
\label{fig:vgg}
\end{figure}

\end{itemize}

\section{Classificação de peixes utilizando CNN}
A base de dados utilizada neste trabalho foi coletada do desafio \textit{The Nature Conservancy Fisheries Monitoring}, da plataforma \textit{Kaggle}. \textit{The Nature Conservancy} é uma organização internacional, sem fins lucrativos, líder na conservação da biodiversidade e do meio ambiente, cuja missão é conservar plantas, animais e comunidades naturais que representam a diversidade da vida na Terra, protegendo espaços que necessitam para sobreviver. De acordo com a organização, cerca de 60\% do atum pescado no mundo é realizado de forma ilegal e não reportado. Essa prática prejudica o ecossistema marinho e os suprimentos de frutos do mar. Com isso, a organização está monitorando essas atividades por meio de imagens, que apesar de funcionar bem, a quantidade de dados produzido é muito grande para serem classificados manualmente. Sendo assim, a organização propôs o desafio para comunidade Kaggle desenvolver algoritmos capazes de detectar e classificar espécies de peixes a partir de imagens obtidas por câmeras instaladas em barcos.  

\begin{figure}[h]
\centering
\includegraphics[scale=0.45]{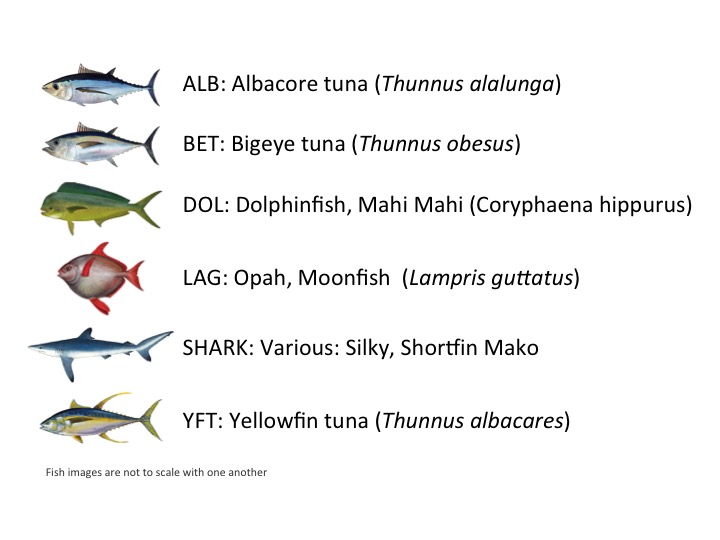}
\caption{Espécies de peixes contidas no desafio em escala ilustrativa}
\label{fig:classpeixes}
\end{figure}

Existem oito categorias a serem classificadas, as seis espécies de peixes ilustradas na Figura \ref{fig:classpeixes} além da classe \textit{outro}, indicando uma espécio de peixe que não seja a da figura, e uma outra classe \textit{no fish}, indicando que não existe nenhum peixe na imagem.

\begin{figure*}
\centering    
\subfigure[Exemplo da espécie ALB]{
\includegraphics[width=0.6\columnwidth]{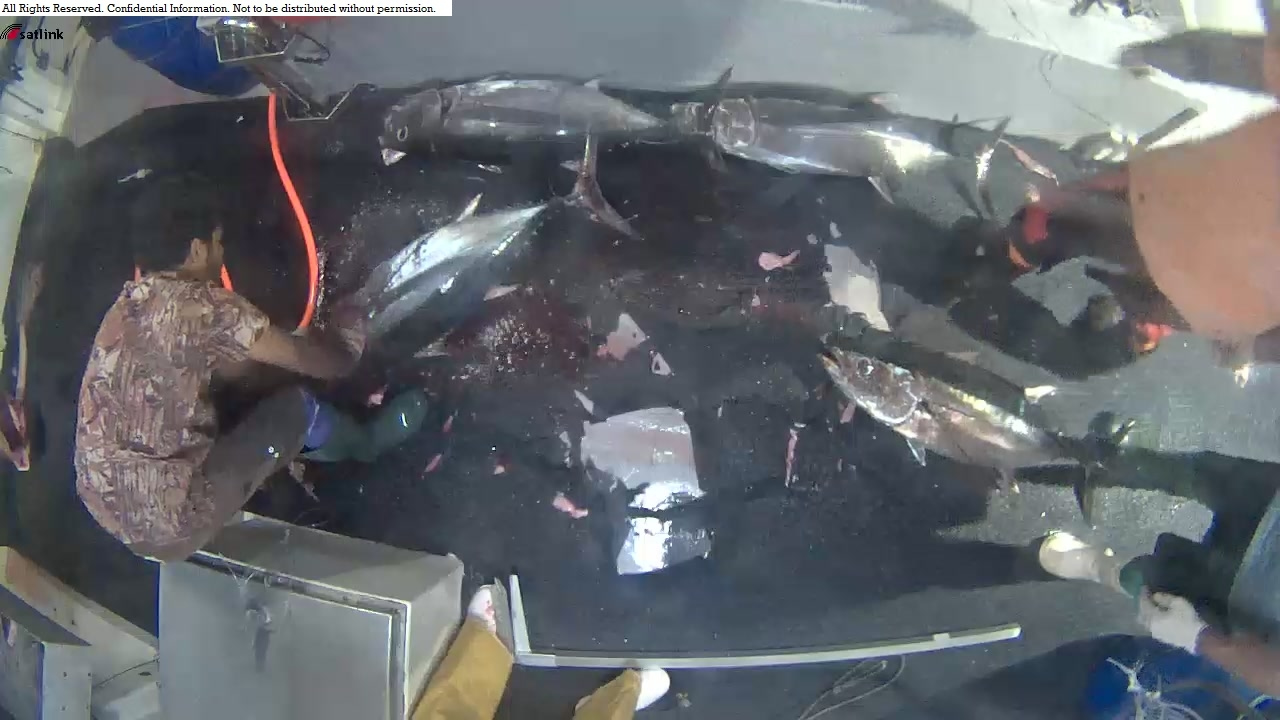}
\label{fig:exemplos_1}
}
\quad
\subfigure[Exemplo da espécie SHARK]{
\includegraphics[width=0.6\columnwidth]{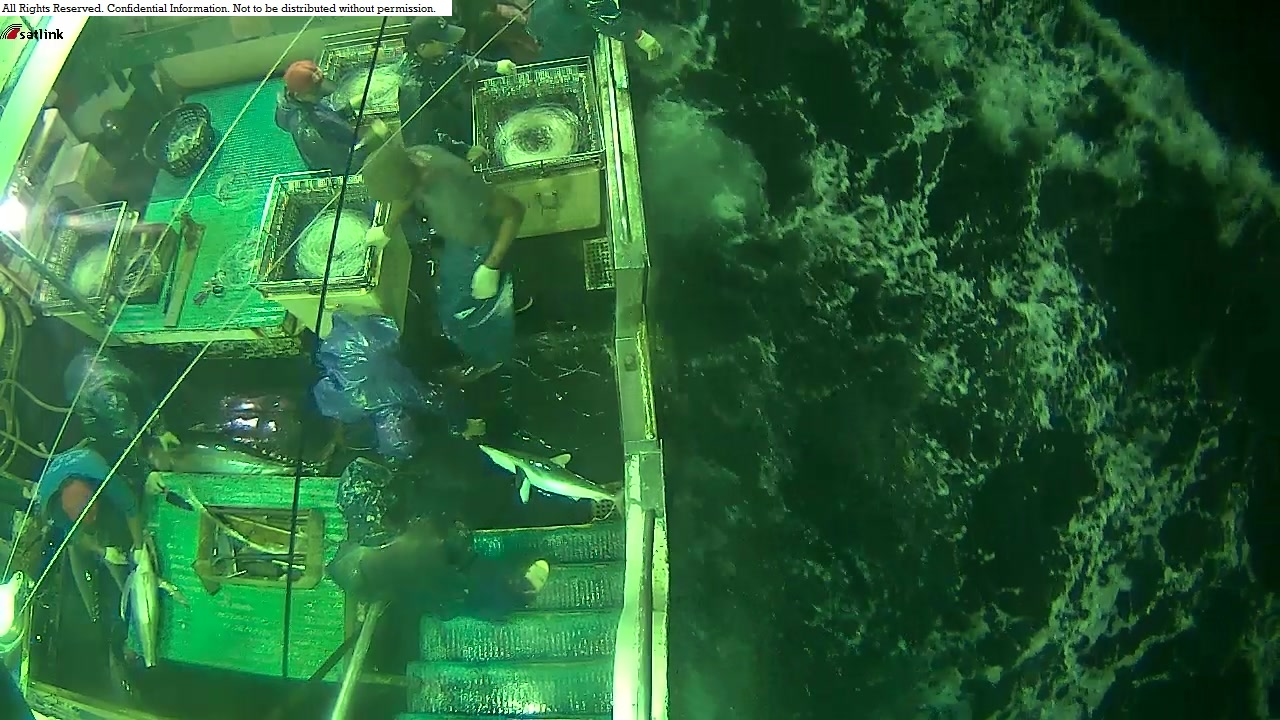}
\label{fig:exemplos_2}
}
\quad
\subfigure[Exemplo da espécie LAG]{
\includegraphics[width=0.6\columnwidth]{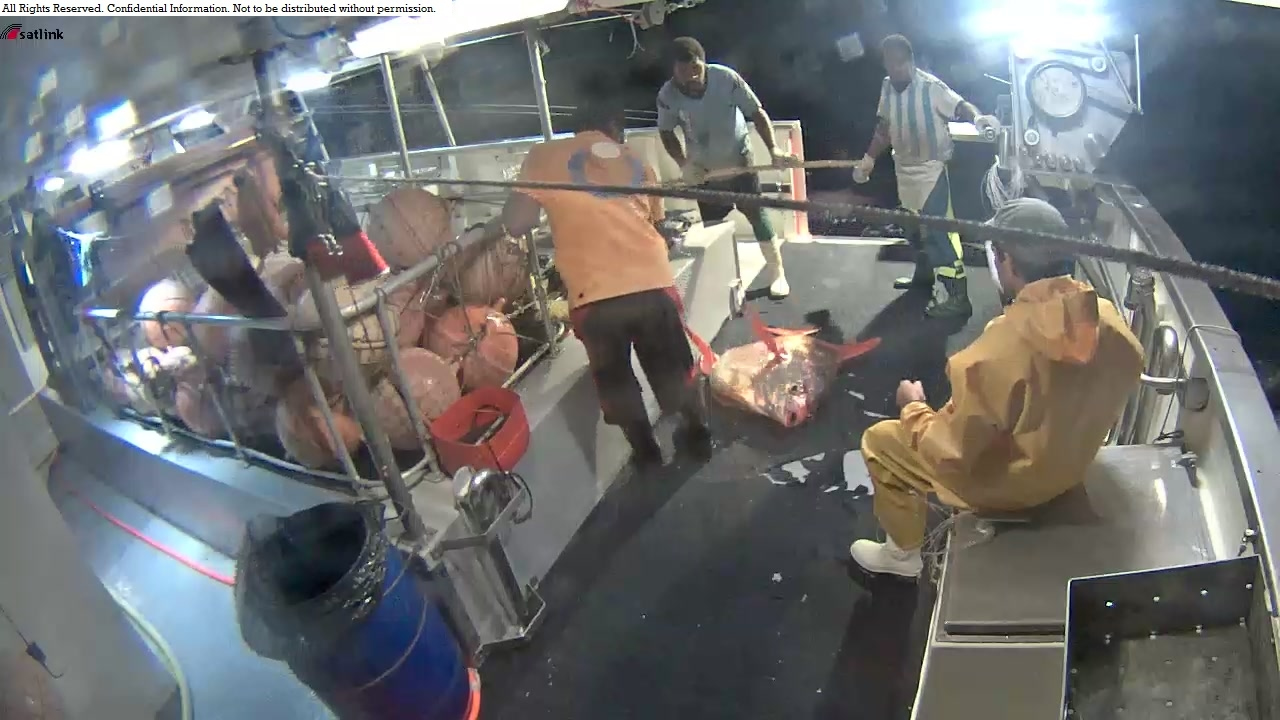}
\label{fig:exemplos_3}

}

\caption{Exemplos de imagens da base dados para classificação de espécies de peixes}
\label{fig:exemplos}
\end{figure*}

\noindent Cada imagem da base de dados possui $1280 \times 974$ \textit{pixels} e apenas uma classe como resposta. O algoritmo deve retornar oito saídas com a probabilidade da amostra de entrada pertencer a cada classe. Obviamente, a classe com maior probabilidade deve ser a escolhida.
Na Figura \ref{fig:exemplos} são ilustradas exemplos de imagens utilizadas no desafio. Como pode ser observado, as imagens possuem diferentes ângulos, iluminação e algumas possuem pequenos peixes, utilizados como isca, que não podem ser confundidos com as espécies desejadas. A quantidade de amostras para cada classe é descrita na Tabela \ref{tab:basepeixes}, o que mostra que a base de dados é bastante desbalanceada, tendo a classe ALB com 1719 amostras e a classe LAG com apenas 67. 

\begin{table}[h]
\centering
\normalsize
\begin{tabular}{cc}
\hline
\textbf{Espécie} & \textbf{Nº de amostras} \\ \hline
ALB              & 1719                    \\
BET              & 200                     \\
DOL              & 117                     \\
LAG              & 67                      \\
NoF              & 465                     \\
Outros           & 299                     \\
Shark            & 176                     \\
YFT              & 734                     \\ \hline
\textbf{Total}   & 3777                    \\ \hline
\end{tabular}
\caption{Número de amostras por espécie contidas na base de dados}
\label{tab:basepeixes}
\end{table}

\section{Experimentos}
Para atacar o problema de classificação de espécie de peixes, neste trabalho foi utilizado uma rede neural convolucional com arquitetura que será descrita a seguir. Para implementar a rede foi utilizada a linguagem \textit{Python} com as bibliotecas \textit{TensorFlow} e \textit{Keras}. A \textit{TensorFLow} permite utilizar GPUs de maneira rápida e fácil e a \textit{Keras} disponibiliza uma série de funções para construção de modelos neurais utilizando a \textit{TensorFlow} para interface com GPUs. Como a \textit{TensorFlow} é baseada em CUDA, para utilizá-la é necessário um \textit{hardware} da NVIDIA.

Para realizar a classificação da base de dados foram testadas diferentes arquiteturas de CNN. A arquitetura que obteve melhor desempenho, e consequentemente foi escolhida para ser utilizada neste trabalho, está ilustrado na Figura \ref{fig:arq-peixes}. Como pode ser observado, o modelo é semelhante à AlexNet, todavia a rede possui quatro camadas de convolução, realizando \textit{maxpooling} a cada duas camadas, duas camadas totalmente conectadas e uma camada de saída \textit{softmax}. Baseado nesta arquitetura foram escolhidos 3 modelos que obtiveram os melhores resultados nos testes. São eles:

\begin{figure}[h]
\centering
\includegraphics[scale=0.25]{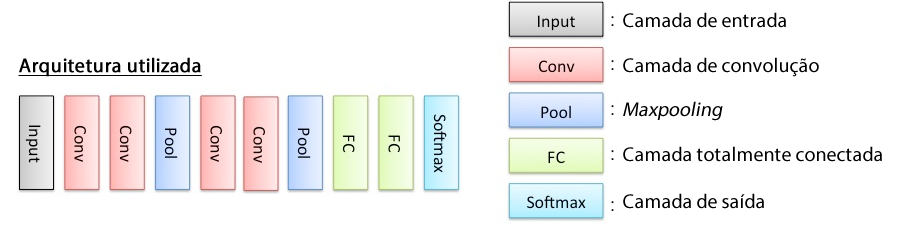}
\caption{Arquitetura utilizada para classificação das espécies de peixe}
\label{fig:arq-peixes}
\end{figure}

\begin{itemize}
\item \textbf{Modelo 1:} utiliza os 4 filtros $3 \times 3$ na Conv$_1$ e Conv$_2$, 8 filtros filtros $3 \times 3$ na conv$_3$ e conv$_4$, 96 neurônios na FC$_1$, 16 na FC$_2$ e 8 na camada de saída.

\item \textbf{Modelo 2:} utiliza os 16 filtros $2 \times 2$ na Conv$_1$ e Conv$_2$, 32 filtros filtros $3 \times 3$ na conv$_3$ e conv$_4$, 96 neurônios na FC$_1$, 16 na FC$_2$ e 8 na camada de saída.

\item \textbf{Modelo 1:} utiliza os 4 filtros $3 \times 3$ na Conv$_1$ e Conv$_2$, 8 filtros filtros $3 \times 3$ na conv$_3$ e conv$_4$, 144 neurônios na FC$_1$, 32 na FC$_2$ e 8 na camada de saída.
\end{itemize}

\noindent Em todos os modelos são utilizados \textit{stride = 1} para as camadas convolucionais. Além disso, as mesmas possuem funções de ativação do tipo ReLU; Todos os \textit{maxpoolings} são realizados com filtros $2 \times 2$ e com \textit{stride = 2}, tanto na horizontal, quanto na vertical; O treinamento é realizado via SDG (\textit{stochastic gradient descent}), com taxa de aprendizado igual $0.01$, \textit{weight decay} igual a $1 \times 10^{-6}$. \textit{momentum} igual a $0.8$, \textit{dropout} com probabilidade igual a 0.5 e \textit{mini-batch} igual a 24. Além disso, é utilizada a técnica de validação cruzada \textit{k-folder} contendo 5 \textit{folders}. Os modelos são treinados utilizando 100 épocas em cada \textit{folder}. Todos esses valores foram obtidos de maneira empírica. Por fim, todos os modelos recebe imagens de $48 \times 48$ \textit{pixels} em RGB.

A métrica de desempenho utilizada pelo \textit{Kaggle} é a \textit{multi-class log loss}, definida como:

\begin{equation}
logloss = -\frac{1}{N} \sum_{i=1}^N \sum_{j=1}^M y_{ij} ln(p_{ij}) 
\end{equation}

\noindent na qual $N$ é o número de amostras, $M$ é o número de classes do problema, $y_{ij}$ é 1 se a amostra $i$ pertence a classe $j$, caso contrário é $0$, e $p_{ij}$ é a probabilidade obtida pelo modelo da amostra $i$ pertencer a classe $j$. Quanto menor o valor obtido pelo $logloss$, melhor é o modelo. Portanto, essa métrica pune classificadores que erra com muita certeza e beneficia aqueles que acertam com muita certeza. 

Como descrito na seção anterior, a base de dados utilizada é bem desbalanceada. Isso é um problema para o treinamento de redes neurais em geral. Além disso, é de conhecimento dos organizadores e dos competidores que se houvesse mais imagens na base, facilitaria muito o trabalho de uma CNN. Todavia, essas restrições fazem parte do problema e fica a cargo dos competidores solucioná-los. Para auxiliar nestes problemas, foi retirado um conjunto de 150 imagens para teste, balanceando as amostras com o número de classes, e as imagens restantes foram replicadas aumentando o número de imagens das classes BET, DOL, LAG, Outros e Shark. Depois desta atualização, cada classe ficou com cerca de 400 imagens. 

Na Tabela \ref{tab:res1} são descritos os resultados de cada modelo. Primeiramente o modelo 1 é testado para a base de dados sem o balanceamento. Na sequência, todos os modelos são executados para a base com balanceamento. A base de validação contida na tabela é constituída das 150 imagens que foram separadas anteriormente e a base de teste é a disponibilizada pelo \textit{Kaggle}, no qual não são conhecidos as respostas para cada amostra, somente o resultado do \textit{logloss} para as 1000 amostras que as contém. As posições no raking do \textit{Kaggle} descritas na tabela são referentes ao dia 12 de dezembro de 2016, como a competição se encerra em 4 meses, novos competidores estão submetendo seus resultados diariamente e esse ranking pode mudar. Vale a pena ressaltar, que até a presente data existem 689 submissões de time ou competidores individuais no desafio.

\begin{table*}[h]
\centering
\begin{tabular}{cccclllll}
\cline{1-5}
\multicolumn{5}{c}{\textbf{Base desbalanceada}}                                                                                 &  &  &  &  \\ \cline{1-5}
\textbf{Modelos}     & \textbf{Logloss treino} & \textbf{Logloss validação} & \textbf{Logloss Kaggle} & \textbf{Pos. Kaggle} &  &  &  &  \\ \cline{1-5}
\textit{Modelo 1}    &0.4012                   &0.9817                            &1.6120                         & \multicolumn{1}{c}{512} &  &  &  &  \\ \cline{1-5}

\multicolumn{5}{c}{\textbf{Base balanceada}}                                                                                    &  &  &  &  \\ \cline{1-5}
\textbf{Modelos}     & \textbf{Logloss treino} & \textbf{Logloss validação} & \textbf{Logloss Kaggle} & \textbf{Pos. Kaggle} &  &  &  &  \\ \cline{1-5}
\textit{Modelo 1}    & 0.3055                  & 0.7080                     & 1.1716                  & \multicolumn{1}{c}{167} &  &  &  &  \\
\textit{Modelo 2}    & 0.2644                  & 0.8012                     & 1.2628                  & \multicolumn{1}{c}{318} &  &  &  &  \\
\textit{Modelo 3}    & 0.2124                  & 0.7618                     & 1.2437                   & \multicolumn{1}{c}{287} &  &  &  &  \\ \cline{1-5}
\multicolumn{1}{l}{} & \multicolumn{1}{l}{}    & \multicolumn{1}{l}{}       & \multicolumn{1}{l}{}    &                         &  &  &  & 
\end{tabular}

\caption{Resultado da classificação da base de dados para cada um dos modelos}
	
\label{tab:res1}
\end{table*}

Como pode ser observado na Tabela \ref{tab:res1}, o balanceamento das imagens contribuiu para melhorar o desempenho das mesmas. Além disso, o modelo 1 foi o que obteve melhor resultado em relação aos demais modelos. Ao aumentar o número de neurônios da camada totalmente conectada a rede começa a sofrer de \textit{overfitting}, começando a decorar os dados de treinamento diminuindo o \textit{logloss} destes dados, mas aumentando o dos testes. De fato, este é um problema esperado, levando em consideração que a base não contém um acervo muito grande de imagens. 

Na Tabela \ref{tab:res2} é exibido uma matriz de confusão para o modelo 1 com a base balanceada. É possível notar que as maiores confusões de classe ocorrem entre a classe ALB e as demais. Isso ocorre, por que mesmo após o balanceamento, a classe ALB continua com muito mais imagens do que as demais.

\begin{table*}[h]
\centering
\begin{tabular}{ccccccccc}
                                     & \textbf{ALB} & \textbf{BET} & \textbf{DOL} & \textbf{LAG} & \textbf{NoF} & \textbf{Outros} & \textbf{Shark} & \textbf{YET} \\ \hline
\multicolumn{1}{c|}{\textbf{ALB}}    & 25           & 0            & 0            & 0            & 0            & 0               & 0              & 0            \\
\multicolumn{1}{c|}{\textbf{BET}}    & 1            & 15           & 0            & 0            & 1            & 0               & 0              & 3            \\
\multicolumn{1}{c|}{\textbf{DOL}}    & 0            & 0            & 15           & 0            & 0            & 0               & 0              & 0            \\
\multicolumn{1}{c|}{\textbf{LAG}}    & 0            & 0            & 0            & 10           & 0            & 0               & 0              & 0            \\
\multicolumn{1}{c|}{\textbf{NoF}}    & 5            & 0            & 0            & 0            & 15           & 0               & 0              & 0            \\
\multicolumn{1}{c|}{\textbf{Outros}} & 2            & 0            & 0            & 0            & 0            & 18              & 0              & 0            \\
\multicolumn{1}{c|}{\textbf{Shark}}  & 2            & 0            & 0            & 0            & 0            & 0               & 18             & 0            \\
\multicolumn{1}{c|}{\textbf{YFT}}    & 2            & 0            & 0            & 0            & 0            & 0               & 0              & 18           \\ \hline
\end{tabular}
\caption{Matriz de confusão para o modelo 1 utilizando a base de dados balanceada}
\label{tab:res2}
\end{table*}

\section{Conclusão}
Neste trabalho uma rede neural convolucional foi utilizada pra classificar espécies de peixes disponibilizados no desafio \textit{The Nature Conservancy Fisheries Monitoring}, da plataforma Kaggle. A base de dados é desafiadora, devido as características das imagens, que possuem ângulos diferentes, muito ruído, iluminação muito diferentes, etc; além da base ser muito desbalanceada e não contemplar um número muito grande de imagens. A arquitetura utilizada para atacar o problema, semelhante à AlexNet, foi dividida em três modelos. Seus resultados foram descritos e discutidos. O melhor modelo obteve posição 167ª dentre as 689 possíveis. O resultado obtido é razoável mas pode ser muito melhorado. Sendo assim, como trabalho futuro, pretende-se melhorar o balanceamento da base de dados, pois os resultados aqui apresentados mostram que isso é essencial para o desempenho da rede; utilizar um modelo pré-treinado com outras bases de dados para ensinar a rede o que é um peixe e em seguida, aplicá-lo para esta base de dados; realizar uma rede modulada, no qual primeiro ela classifica se o peixe é da classe ALB ou nao, e na sequência realiza a classificação para as demais classes; e por fim, utilizar um \textit{ensemble} com outros classificadores para melhorar o desempenho da rede. Tendo em vista que a competição se encerra daqui a quatro meses, todos esses pontos podem ser implementados a fim de melhorar a classificação no ranking final.

\section*{Referências}
\bibliography{references}

\end{document}